\pdfoutput=1
\documentclass[10pt,twocolumn,letterpaper]{article}

\PassOptionsToPackage{table}{xcolor}
\usepackage[pagenumbers]{paperstyle}

\makeatletter
\patchcmd{\@maketitle}{\vskip .375in}{\vskip 0in}{}{\PackageWarning{arxiv}{title space patch 1 failed}}
\patchcmd{\@maketitle}{\vspace*{12pt}}{\vspace*{0pt}}{}{\PackageWarning{arxiv}{title space patch 3 failed}}
\patchcmd{\@maketitle}{\vspace*{24pt}}{\vspace*{7pt}}{}{\PackageWarning{arxiv}{title space patch 2 failed}}
\patchcmd{\@maketitle}{\vskip .5em}{\vskip .3em}{}{\PackageWarning{arxiv}{title space patch 4 failed}}
\patchcmd{\abstract}{\vspace*{12pt}}{\vspace*{4pt}}{}{\PackageWarning{arxiv}{abstract space patch failed}}
\patchcmd{\endabstract}{\vspace*{12pt}}{\vspace*{6pt}}{}{\PackageWarning{arxiv}{endabstract space patch failed}}
\makeatother
\usepackage[T1]{fontenc}
\usepackage{multirow}
\usepackage{algorithm}
\usepackage{algpseudocode}
\usepackage{tikz}
\usepackage[most]{tcolorbox}
\usepackage{amsmath,amssymb}
\usepackage{xcolor}
\usepackage{algorithm}
\usepackage{algpseudocode}

\definecolor{bosGray}{HTML}{6B7280}

\algrenewcommand{\algorithmicrequire}{\textbf{Input:}}
\algrenewcommand{\algorithmicensure}{\textbf{Output:}}
\algrenewcommand{\algorithmiccomment}[1]{%
  \hfill{\color{bosGray}$\triangleright$ #1}%
}

\usepackage{xcolor}
\usepackage[most]{tcolorbox}
\tcbuselibrary{breakable,listings,skins}

\definecolor{mmstarColor}{HTML}{1D4ED8}
\definecolor{m3cotColor}{HTML}{0F766E}
\definecolor{mmbenchColor}{HTML}{7E22CE}
\definecolor{popeColor}{HTML}{C2410C}
\definecolor{raceColor}{HTML}{15803D}
\definecolor{logiqaColor}{HTML}{BE123C}
\definecolor{mathqaColor}{HTML}{0369A1}
\definecolor{quacColor}{HTML}{7C3E1D}

\newtcblisting{datasetprompt}[2]{
  enhanced,
  breakable,
  listing only,
  title={#1},
  colback=#2!6!white,
  colframe=#2,
  colbacktitle=#2,
  coltitle=white,
  coltext=black,
  fonttitle=\small\bfseries\sffamily,
  arc=1.5mm,
  outer arc=1.5mm,
  boxrule=0.7pt,
  titlerule=0pt,
  left=2.5mm,
  right=2.5mm,
  top=1.8mm,
  bottom=1.8mm,
  before skip=0.65em,
  after skip=0.8em,
  listing options={
    basicstyle=\footnotesize\ttfamily,
    columns=fullflexible,
    keepspaces=true,
    showstringspaces=false,
    breaklines=true,
    breakatwhitespace=true,
    aboveskip=0pt,
    belowskip=0pt
  }
}

\newcommand{\blackcircled}[1]{%
  \tikz[baseline=(char.base)]{
    \node[
      circle,
      fill=black,
      text=white,
      minimum size=1.1em,
      inner sep=0pt,
      font=\sffamily\bfseries\scriptsize
    ] (char) {#1};
  }%
}

\newtcolorbox{keyquestion}{
    enhanced,
    colback=gray!5,
    colframe=black!50,
    boxrule=0.6pt,
    arc=2pt,
    left=6pt,
    right=6pt,
    top=7pt,
    bottom=5pt,
    before skip=8pt,
    after skip=6pt,
    title={Key Question},
    coltitle=black,
    fonttitle=\bfseries,
    attach boxed title to top left={
        xshift=6pt,
        yshift=-3pt
    },
    boxed title style={
        colback=gray!15,
        colframe=black!50,
        boxrule=0.6pt,
        arc=2pt,
        left=4pt,
        right=4pt,
        top=2pt,
        bottom=2pt
    }
}

\definecolor{linkblue}{rgb}{0.21,0.49,0.74}
\usepackage[pagebackref,breaklinks,colorlinks,allcolors=linkblue]{hyperref}

\title{BOS Summary: The First Token Matters}
\author{Hui Zhu$^{1}$ \quad Saleh Zare Zade$^{1}$ \quad Rafi Ibn Sultan$^{1}$\\
Deng Pan$^{2}$ \quad Arun Ross$^{3}$ \quad Dongxiao Zhu$^{1}$\\[4pt]
$^{1}$Wayne State University \quad $^{2}$University of Notre Dame \quad $^{3}$Michigan State University\\
{\tt\small \{hq2197, hr3331, hm4013, ct4442\}@wayne.edu, dpan@nd.edu, rossarun@msu.edu}}
\begin{document}
\maketitle

\begin{abstract}
Large vision-language models (LVLMs) and language models (LLMs) often exhibit
\emph{attention sinks}, whereby particular positions, such as the
beginning-of-sequence (BOS) token, receive a disproportionate share of
attention. Under standard causal masking, however, the BOS query attends only
to itself, so the representation that later tokens repeatedly access carries
no information about the input. We introduce \textbf{BOS Summary}, a
training-free inference-time method that repurposes this persistent pathway
to carry input context. At a single transformer layer during prefill, BOS
Summary relaxes the causal constraint for the BOS query row alone, letting
BOS aggregate all visual and textual input tokens into a self-derived context
representation while every other token retains standard causal attention. The
enriched representation then remains accessible to the model throughout
decoding, with no additional parameters, auxiliary modules, or decoding-time
overhead. Across multiple VLM and LLM families and eight multimodal and
text-only benchmarks, BOS Summary improves over direct and Chain-of-Thought
prompting and outperforms recent sink-intervention methods on multi-step
reasoning tasks. A mechanistic analysis shows that making BOS input-specific
also weakens its sink dominance and redistributes attention toward
content-bearing prompt tokens, indicating that changing what a sink carries
reshapes how later layers allocate attention.
\end{abstract}

\section{Introduction}

Recent advances in vision-language models (VLMs) have enabled strong
performance across multimodal understanding, visual reasoning, and
instruction-following tasks~\citep{liu2023llava,bai2025qwen25vl}. Despite this
progress, VLMs remain prone to hallucination and context loss: during
long-form generation, they may produce statements unsupported by the image
or gradually deviate from the original instruction. These limitations are
well documented by hallucination benchmarks
~\citep{li2023evaluating,wang2023amber,rohrbach2018chair} and have motivated both
training-free decoding methods~\citep{huang2024opera,leng2024mitigating,chuang2024dola}
and instruction-tuning approaches~\citep{liu2024mitigating}. One mechanism associated
with these failures is \emph{attention drift}~\citep{liu2026sinktrack,huang2024opera}:
as generation proceeds, attention can shift toward previously generated tokens
and away from the grounding image and input instruction. This problem becomes
particularly important in extended multimodal generation
~\citep{chen2024m3cot}, where visual evidence must remain accessible across
many decoding steps even as the generated context continues to grow.

This motivates us to look for a representation that remains available
throughout decoding and is already repeatedly accessed by later tokens.
Transformers naturally provide such a candidate through the
\emph{attention-sink} phenomenon, in which the first token receives a
disproportionate share of attention from subsequent tokens, often largely
independent of its content~\citep{xiao2024streaming,barbero2025first,
sun2024massive,gu2025attention}. Related sink behavior has also been observed
in vision and vision-language models~\citep{darcet2024registers,
kang2025see}, and recent work suggests that sink representations can
influence subsequent computation rather than merely absorbing attention
~\citep{yoo2026natureattentionsinkshapes}. In our setting, this first position is the
beginning-of-sequence (BOS) token. Because later tokens repeatedly access BOS
throughout decoding, conditioning its representation on the visual input and
instruction could turn this existing pathway into a persistent carrier of
prompt context.

\begin{keyquestion}
Can we turn the first token from an uninformative attention sink into a
context-dependent anchor that gathers information from the visual input
and instruction, allowing the model to access this information throughout
generation?
\end{keyquestion}

Recent inference-time methods demonstrate that attention sinks can serve as useful
control points for shaping model behavior, either by rescaling the attention they
receive or by injecting additional information into their representations
~\citep{han2025zerotuning,liu2026sinktrack}. Rather than externally modifying the sink
or its incoming attention, we explore whether BOS can derive a prompt representation
directly through the model's own attention. This offers a complementary way to exploit
the sink: changing \emph{what BOS carries} using information already present in the
prompt.

\begin{figure}[t]
    \centering
    \includegraphics[width=\columnwidth]{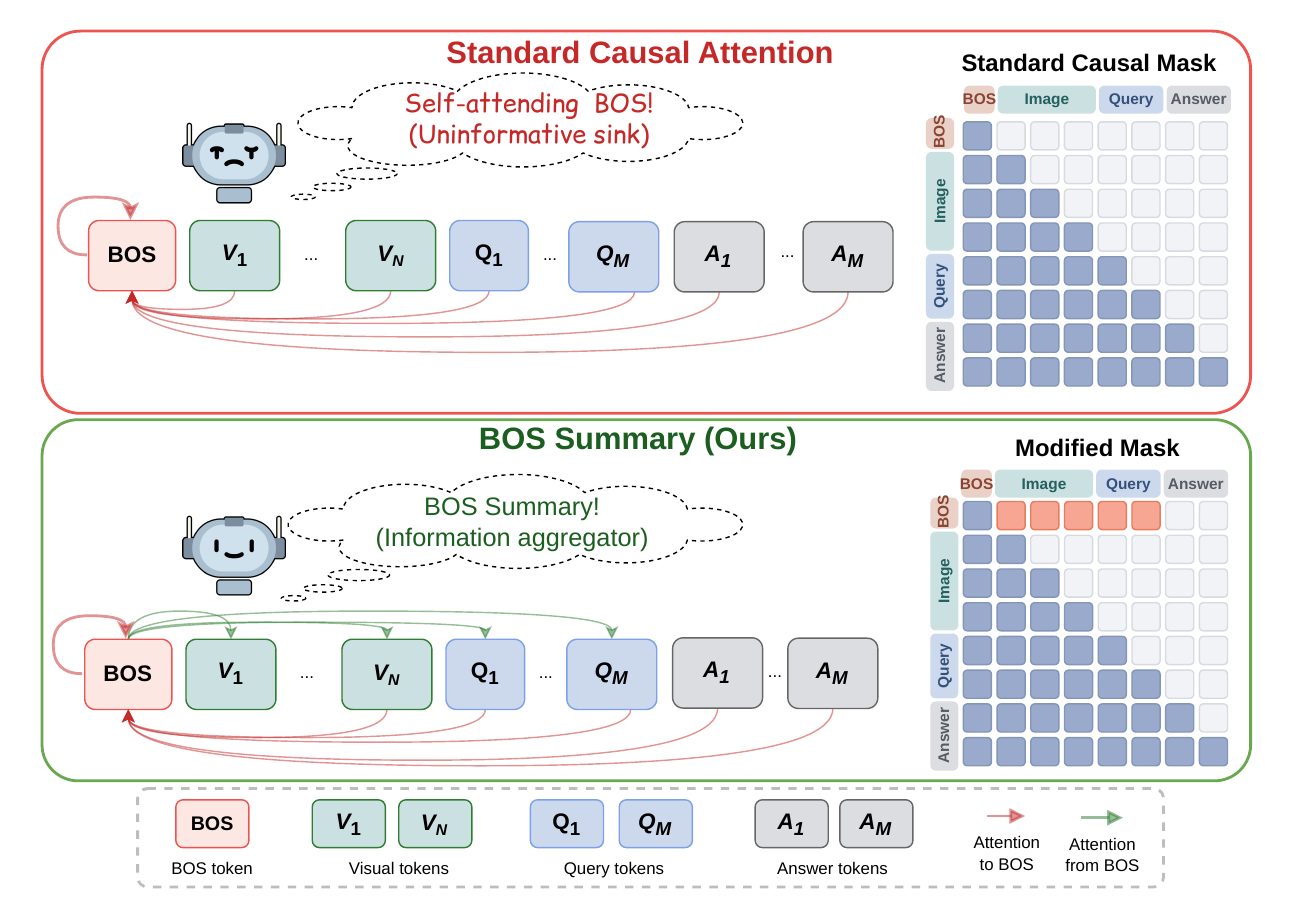}
    \caption{BOS Summary. Standard causal attention leaves BOS attending only to itself, making it an uninformative sink despite receiving heavy attention from all later tokens (top). We unmask the BOS row at one selected layer so BOS attends to the visual and instruction tokens, turning it into a context aggregator that later tokens read from throughout generation (bottom). All other attention patterns, the architecture are unchanged.}
    \label{fig:overview}
\end{figure}

As illustrated in Figure~\ref{fig:overview}, \textbf{BOS Summary} turns the existing
BOS token into a training-free, one-token summary of the visual input and instruction.
Under standard causal attention, BOS attends only to itself, while subsequent tokens
can attend back to BOS throughout generation. We relax the causal constraint only for
the BOS query row at a selected layer during prefill, allowing BOS to aggregate the
visual and textual prompt while leaving the rest of the causal attention mask unchanged.
This simple intervention repurposes an existing attention pathway into a persistent
carrier of prompt context without changing the model architecture or decoding procedure.

Building on this design, our contributions are threefold:
\begin{itemize}
    \item We introduce \textbf{BOS Summary}, a training-free, plug-and-play
    intervention that turns BOS into a one-token, prompt-conditioned
    representation via a single prefill-time mask edit, with no retraining,
    auxiliary modules, or decoding-time cost.
    \item We show that BOS Summary \textbf{improves reasoning and grounding}
    across diverse VLMs, LLMs, and benchmarks targeting context forgetting
    and object hallucination, where faithful access to the input matters most.
    \item We provide a mechanistic analysis showing that an input-specific BOS
    weakens the sink's attention dominance and shifts attention back toward
    content-bearing input tokens, revealing that \textbf{changing what a sink
    carries reshapes how later layers allocate attention}.
\end{itemize}

\section{Related Work}
\label{sec:related}

\paragraph{Attention Sink.}
Transformer language models often concentrate disproportionate,
content-independent attention on the first token, forming an
\emph{attention sink}~\cite{xiao2024streaming}; related sinks also appear in
VLMs on structural and weakly informative visual tokens~\cite{kang2025see, zhang2026drives}.
Prior work studies their mechanisms~\cite{sun2024massive,cancedda2024spectral,
qiu2026unified,guo2024activedormant}, emergence~\cite{
gu2025attention,ruscio2026you}, and functional role~\cite{barbero2025first,zhang2026attention},
with analogous phenomena motivating register tokens in vision transformers
~\cite{darcet2024registers} and finer VLM sink analyses~\cite{choi2026sinks}.
Recent work further suggests that sink representations can carry global
information and shape decoding~\cite{yoo2026natureattentionsinkshapes}, motivating inference-time
interventions~\cite{han2025zerotuning,yu2024unveiling,liu2026sinktrack,liu2026sinkrouter}.
We build on this view by asking what the sink can be made to carry and how a
minimal intervention changes its downstream role.

\paragraph{Repurposing the sink at inference time.}
A growing body of training-free methods intervenes on the attention sink to shape model
behavior without modifying model weights, differing primarily in what aspect of the sink
they modify. The first group changes \emph{how much attention the sink receives}:
ZeroTuning~\cite{han2025zerotuning} calibrates per-head attention to the initial token,
and ACT~\cite{yu2024unveiling} rescales attention for activation steering. The second group
changes \emph{what the sink carries}: SinkTrack~\cite{liu2026sinktrack} injects externally
derived context features, and learned gating can regulate the information stored in the
sink~\cite{choi2026sinks}. Closest to our work, OutRo~\cite{yoo2026natureattentionsinkshapes} also relaxes the
causal mask on sink positions, but couples it with sink detection and head-output
rotation applied throughout decoding, and reports that the mask relaxation alone yields
only marginal gains on video and audio-visual QA. A third group treats sink dominance as
an \emph{efficiency signal} for skipping saturated attention heads~\cite{liu2026sinkrouter}.
BOS Summary isolates mask relaxation as a prefill-only edit applied solely to the BOS query row, while leaving the causal attention mask for all other query rows and the decoding process unchanged. Without injection, gating, detection, rotation, or added parameters, it shows that this minimal intervention is effective for both image-grounded and text-only reasoning across several models and benchmarks.

\paragraph{Attention control for grounding and hallucination.}
Beyond sink-specific interventions, training-free methods mitigate VLM
hallucination through attention and decoding. Attention-based approaches
reweight visual tokens or suppress non-informative attention
~\cite{liu2024paying,yin2025clearsight,arif2025paint,zhang2023tell,gong2024damro},
while decoding methods contrast logits across visual inputs,
depth, or image-biased distributions~\cite{leng2024mitigating,chuang2024dola,
chen2024halc,wang2024mitigating}, or use attention-aware
beam-search penalties~\cite{huang2024opera}. Other methods revise outputs
\textit{post hoc}~\cite{yin2024woodpecker} or reduce hallucination through
training~\cite{liu2024mitigating}. In contrast, BOS Summary modifies one
sink position and uses its downstream influence rather than reweighting content
tokens, altering logits, or correcting outputs. Further discussion appears in
Appendix~\ref{app:related}.

\section{Methodology}
\label{sec:method}

\paragraph{Motivation.}
During autoregressive generation, the visual input and instruction are represented by a
fixed set of prompt tokens, whereas the decoding history grows at every step. Information
from the original prompt must therefore compete with an increasingly long sequence of
generated tokens. A natural way to preserve access to this information is to encode it
into a fixed representation that remains available throughout decoding. The BOS token is
particularly well suited for this role: it is present from the beginning of the sequence,
remains in the cached context, and often receives substantial attention from later tokens
due to the attention-sink phenomenon~\citep{xiao2024streaming,barbero2025first}.

Under standard causal masking, however, the BOS query can attend only to itself and
therefore cannot incorporate information from the visual and textual tokens that follow
it. This creates an underused pathway: later tokens repeatedly attend to BOS, yet the
representation they access is not conditioned on the prompt. BOS Summary exploits this
asymmetry by relaxing the causal constraint only for the BOS query at a selected layer
during prefill, allowing BOS to aggregate information from the multimodal prompt while
leaving the causal attention pattern of every other token unchanged.

The resulting prompt-conditioned BOS state propagates through subsequent transformer
layers and remains accessible to later generated tokens, providing a persistent pathway
to information from the original prompt through a sequence position that the model
already attends to.

\paragraph{Formal Definition.}
We consider a decoder-only transformer (VLM or LLM) with $L$ layers. Let
$X^{(\ell)} \in \mathbb{R}^{n \times d}$ denote the input hidden states
to layer $\ell$, where $n$ is the sequence length, $d$ is the hidden
dimension, and $X^{(\ell)}_0$ corresponds to the BOS token. For clarity,
we present a single-head formulation and omit architecture-specific
operations (e.g., normalization and positional encoding), since BOS
Summary modifies only the attention mask, and the same mask edit is
applied uniformly across all heads. Queries, keys, and values are
computed as
\begin{equation}
\begin{aligned}
    Q &= X^{(\ell)}W_Q,\\
    K &= X^{(\ell)}W_K,\\
    V &= X^{(\ell)}W_V,
\end{aligned}
\label{eq:qkv}
\end{equation}
where $W_Q,W_K,W_V\in\mathbb{R}^{d\times d_k}$ are the model's frozen
projection matrices. The attention weights and head output are
\begin{equation}
    A=
    \operatorname{softmax}\!\left(
        \frac{QK^\top}{\sqrt{d_k}}+M
    \right),
    \qquad
    O=AV,
\label{eq:attn}
\end{equation}
\begin{algorithm}[!ht]
\caption{BOS Summary}
\label{alg:bos-summary}
\footnotesize
\begin{algorithmic}[1]

\Require Hidden states $X^{(\ell)}\in\mathbb{R}^{n\times d}$,
current layer $\ell$, intervention layer $\ell^\star$,
and prefill indicator $p$
\Ensure Attention output $O$

\Function{BOSSummaryAttention}{$X^{(\ell)},\ell,\ell^\star,p$}
    \State $(Q,K,V) \gets \Call{ProjectQKV}{X^{(\ell)}}$
    \State $M \gets \Call{NativeCausalMask}{n}$
    \If{$p \land \ell = \ell^\star$}
        \For{$j \gets 0$ \textbf{to} $n-1$}
            \State $M_{0,j} \gets 0$
        \EndFor
        \Comment{Allow BOS to attend to the full prompt}
    \EndIf
    \State $A \gets
        \Call{Softmax}{
        \dfrac{QK^{\top}}{\sqrt{d_k}} + M}$
    \State $O \gets AV$
    \State \Return $O$
\EndFunction

\end{algorithmic}
\end{algorithm}

where $M\in\{0,-\infty\}^{n\times n}$ is the additive attention mask
applied before the softmax. Under standard causal masking,
$M^{\mathrm{causal}}_{ij}=0$ for $j\le i$ and $-\infty$ otherwise, so
the BOS query attends only to itself, yielding $A_{0,\cdot}$ one-hot at
position~$0$ and $O_0=V_0$.
BOS Summary applies its intervention at a single layer
$\ell^\star\in\{1,\ldots,L\}$ during the prefill pass, i.e., while the
entire multimodal prompt is encoded before autoregressive decoding.
The mask $M$ equals $M^{\mathrm{causal}}$ at every layer except
$\ell^\star$; at this layer we replace only its first row,
\begin{equation}
    M_{0,j}=0,
    \qquad j=0,\ldots,n-1,
\label{eq:mask_edit}
\end{equation}
while leaving every other row unchanged. Since $M_{0,j}=0$ for all $j$,
the softmax term in Equation~\eqref{eq:attn} reduces at position $0$ to
an unmasked attention distribution, and the BOS attention output becomes
a content-weighted aggregation of the entire prompt,
\begin{equation}
    O_0
    =
    \sum_{j=0}^{n-1}A_{0j}V_j,
    \qquad
    A_{0j}
    =
    \operatorname{softmax}_j
    \!\left(
        \frac{Q_0K^\top}{\sqrt{d_k}}
    \right).
\label{eq:bos_summary}
\end{equation}
The resulting BOS attention output is incorporated into the BOS hidden
state through the transformer's standard residual update. At every layer
above $\ell^\star$, this self-derived summary shapes the BOS key and
value representations, allowing later tokens to access information from
the original prompt through the BOS position; the summary's influence
thus depends on the choice of $\ell^\star$
(see Appendix~\ref{app:hyperparameters}). During autoregressive decoding,
BOS Summary does not modify the attention mask; the model therefore
follows its standard causal attention pattern.
Algorithm~\ref{alg:bos-summary} provides pseudocode for our method,
while Appendix~\ref{app:fig_bos_summary} presents a visual overview of
the procedure.

\newcommand{\meanstd}[2]{\ensuremath{#1_{\pm #2}}}
\newcommand{\bestms}[2]{\ensuremath{\mathbf{#1}_{\pm #2}}}
\newcommand{\secondms}[2]{\ensuremath{\underline{#1}_{\pm #2}}}

\begin{table*}[!t]
    \centering
    \caption{
        Performance of different inference methods across four multimodal
        benchmarks. Per-benchmark results are reported as mean Accuracy
        (\%) and Macro-F1 with standard deviation. The Average columns report
        the arithmetic mean of the benchmark scores and the arithmetic mean
        of their benchmark-wise standard deviations. The best mean within
        each model block is shown in bold, while the second-best mean is
        underlined.
    }
    \label{tab:main_results}
    \resizebox{\textwidth}{!}{%
    \begin{tabular}{lcccccccccc}
        \toprule
        \multirow{2}{*}{Method}
        & \multicolumn{2}{c}{MMStar}
        & \multicolumn{2}{c}{M3CoT}
        & \multicolumn{2}{c}{POPE}
        & \multicolumn{2}{c}{MMBench}
        & \multicolumn{2}{c}{Average} \\
        \cmidrule(lr){2-3}
        \cmidrule(lr){4-5}
        \cmidrule(lr){6-7}
        \cmidrule(lr){8-9}
        \cmidrule(lr){10-11}
        & Acc. & Macro-F1
        & Acc. & Macro-F1
        & Acc. & Macro-F1
        & Acc. & Macro-F1
        & Acc. & Macro-F1 \\
        \midrule

        \rowcolor{gray!8}
        \multicolumn{11}{c}{
            \textit{Qwen2.5-VL-7B-Instruct}
        } \\
        \midrule

        Direct
        & \meanstd{52.00}{6.40}
        & \meanstd{68.16}{5.87}
        & \meanstd{47.24}{2.82}
        & \meanstd{62.22}{3.97}
        & \meanstd{83.53}{1.40}
        & \meanstd{83.19}{1.48}
        & \meanstd{85.03}{1.42}
        & \meanstd{91.93}{0.87}
        & \meanstd{66.95}{1.78}
        & \meanstd{76.38}{1.92} \\

        CoT
        & \meanstd{58.60}{4.61}
        & \meanstd{73.77}{3.43}
        & \meanstd{62.99}{2.33}
        & \bestms{75.87}{4.61}
        & \meanstd{84.91}{1.70}
        & \meanstd{85.07}{1.75}
        & \meanstd{83.99}{1.57}
        & \meanstd{91.27}{0.91}
        & \meanstd{72.62}{1.54}
        & \meanstd{81.50}{1.81} \\

        ZeroTuning
        & \secondms{59.87}{4.36} & \secondms{74.68}{3.44}
        & \secondms{63.72}{1.98} & \secondms{75.21}{2.67}
        & \secondms{85.66}{1.00} & \secondms{85.59}{1.03}
        & \bestms{86.69}{1.00} & \bestms{92.90}{0.58}
        & \secondms{73.98}{1.09} & \secondms{82.09}{1.51} \\

        SinkTrack
        & \secondms{59.87}{2.38}
        & \meanstd{74.56}{2.14}
        & \meanstd{62.68}{2.14}
        & \meanstd{74.89}{3.44}
        & \bestms{86.47}{0.74}
        & \bestms{86.37}{0.78}
        & \secondms{86.44}{0.81}
        & \secondms{92.75}{0.48}
        & \meanstd{73.86}{0.97}
        & \bestms{82.14}{1.15} \\

        BOS Summary
        & \bestms{61.47}{2.77}
        & \bestms{75.55}{2.44}
        & \bestms{65.14}{2.44}
        & \meanstd{74.94}{8.77}
        & \meanstd{85.01}{0.34}
        & \meanstd{85.15}{0.39}
        & \meanstd{85.47}{1.43}
        & \meanstd{92.17}{0.87}
        & \bestms{74.27}{1.03}
        & \meanstd{81.95}{1.84} \\
        \midrule

        \rowcolor{gray!8}
        \multicolumn{11}{c}{
            \textit{Qwen3-VL-8B-Instruct}
        } \\
        \midrule

        Direct
        & \meanstd{48.07}{5.78}
        & \meanstd{64.60}{5.81}
        & \meanstd{31.79}{3.10}
        & \meanstd{46.62}{2.30}
        & \secondms{89.23}{1.13}
        & \secondms{89.20}{1.15}
        & \meanstd{84.08}{1.99}
        & \meanstd{91.32}{1.20}
        & \meanstd{63.29}{1.67}
        & \meanstd{72.94}{1.52} \\

        CoT
        & \secondms{66.67}{3.60}
        & \secondms{79.80}{2.54}
        & \secondms{65.92}{1.48}
        & \secondms{77.36}{3.60}
        & \meanstd{88.08}{0.77}
        & \meanstd{88.09}{0.81}
        & \secondms{88.38}{1.57}
        & \secondms{93.84}{0.85}
        & \secondms{77.26}{1.18}
        & \secondms{84.77}{1.31} \\

        ZeroTuning
        & \meanstd{64.80}{3.15} & \meanstd{78.55}{2.51}
        & \meanstd{59.15}{3.24} & \meanstd{71.09}{5.30}
        & \meanstd{88.29}{0.45} & \meanstd{88.32}{0.41}
        & \meanstd{87.04}{0.81} & \meanstd{93.01}{0.55}
        & \meanstd{74.82}{1.19} & \meanstd{82.74}{0.66} \\

        SinkTrack
        & \meanstd{63.40}{2.97}
        & \meanstd{77.54}{2.43}
        & \meanstd{59.53}{2.16}
        & \meanstd{71.99}{3.63}
        & \meanstd{88.48}{0.43}
        & \meanstd{88.49}{0.44}
        & \meanstd{87.09}{0.88}
        & \meanstd{93.05}{0.48}
        & \meanstd{74.62}{0.91}
        & \meanstd{82.77}{0.98} \\

        BOS Summary
        & \bestms{68.07}{2.28}
        & \bestms{81.05}{1.59}
        & \bestms{68.08}{0.73}
        & \bestms{80.02}{1.12}
        & \bestms{89.29}{0.53}
        & \bestms{89.21}{1.14}
        & \bestms{89.05}{0.85}
        & \bestms{94.19}{0.47}
        & \bestms{78.62}{0.74}
        & \bestms{86.12}{0.86} \\
        \midrule

        \rowcolor{gray!8}
        \multicolumn{11}{c}{
            \textit{Gemma-3-4B-Instruct}
        } \\
        \midrule

        Direct
        & \meanstd{36.80}{2.14}
        & \meanstd{53.25}{2.10}
        & \meanstd{32.14}{1.87}
        & \meanstd{44.64}{2.43}
        & \bestms{83.94}{0.84}
        & \bestms{83.93}{0.85}
        & \meanstd{85.43}{0.69}
        & \secondms{91.35}{1.20}
        & \meanstd{59.58}{0.94}
        & \meanstd{68.29}{1.08} \\

        CoT
        & \secondms{50.67}{4.17}
        & \secondms{67.01}{3.55}
        & \meanstd{49.44}{2.75}
        & \secondms{58.76}{5.79}
        & \meanstd{82.18}{0.98}
        & \meanstd{82.12}{0.97}
        & \secondms{85.54}{2.23}
        & \meanstd{91.28}{2.54}
        & \secondms{66.96}{1.42}
        & \secondms{74.79}{1.67} \\

        ZeroTuning
        & \meanstd{49.20}{2.99} & \meanstd{65.46}{2.64}
        & \meanstd{49.66}{1.67} & \meanstd{57.94}{5.39}
        & \meanstd{82.42}{0.98} & \meanstd{82.39}{0.99}
        & \meanstd{80.94}{1.07} & \meanstd{89.41}{0.59}
        & \meanstd{65.56}{1.57} & \meanstd{73.80}{1.33} \\

        SinkTrack
        & \meanstd{48.53}{5.25}
        & \meanstd{64.90}{4.85}
        & \secondms{50.99}{2.82}
        & \bestms{61.02}{4.02}
        & \meanstd{82.63}{0.78}
        & \meanstd{82.60}{0.78}
        & \meanstd{81.17}{1.25}
        & \meanstd{89.53}{0.67}
        & \meanstd{65.83}{1.23}
        & \meanstd{74.51}{1.31} \\

        BOS Summary
        & \bestms{52.49}{1.97}
        & \bestms{68.39}{1.85}
        & \bestms{51.73}{2.59}
        & \meanstd{58.48}{6.76}
        & \secondms{82.76}{0.69}
        & \secondms{82.76}{0.71}
        & \bestms{86.01}{1.42}
        & \bestms{91.41}{0.95}
        & \bestms{68.25}{1.08}
        & \bestms{75.26}{1.54} \\
        \midrule

        \rowcolor{gray!8}
        \multicolumn{11}{c}{
            \textit{Gemma-3-12B-Instruct}
        } \\
        \midrule

        Direct
        & \meanstd{45.60}{4.06}
        & \meanstd{61.69}{4.06}
        & \meanstd{45.38}{2.92}
        & \meanstd{56.17}{6.48}
        & \meanstd{84.57}{1.15}
        & \meanstd{84.55}{1.16}
        & \meanstd{85.44}{1.47}
        & \meanstd{91.37}{1.89}
        & \meanstd{65.25}{1.49}
        & \meanstd{73.44}{1.73} \\

        CoT
        & \secondms{60.00}{4.70}
        & \secondms{74.68}{3.93}
        & \secondms{59.62}{2.90}
        & \secondms{69.56}{4.20}
        & \bestms{85.14}{0.59}
        & \bestms{85.13}{0.61}
        & \secondms{86.00}{1.55}
        & \bestms{91.74}{1.78}
        & \secondms{72.69}{1.36}
        & \secondms{80.28}{1.51} \\

        ZeroTuning
        & \meanstd{58.40}{2.05} & \meanstd{73.08}{2.03}
        & \meanstd{58.37}{3.96} & \meanstd{69.70}{4.96}
        & \meanstd{84.69}{0.20} & \meanstd{84.76}{0.23}
        & \meanstd{84.50}{1.34} & \secondms{91.58}{0.73}
        & \meanstd{71.49}{1.16} & \meanstd{79.78}{0.21} \\

        SinkTrack
        & \meanstd{57.80}{1.85}
        & \meanstd{72.48}{1.89}
        & \meanstd{58.07}{3.10}
        & \meanstd{68.16}{4.44}
        & \meanstd{84.70}{0.36}
        & \secondms{84.80}{0.36}
        & \meanstd{84.34}{1.23}
        & \meanstd{91.48}{0.70}
        & \meanstd{71.23}{1.02}
        & \meanstd{79.23}{1.17} \\

        BOS Summary
        & \bestms{62.58}{1.35}
        & \bestms{76.43}{0.88}
        & \bestms{61.06}{1.10}
        & \bestms{72.25}{4.08}
        & \secondms{84.73}{1.01}
        & \meanstd{84.73}{1.03}
        & \bestms{87.13}{0.86}
        & \meanstd{91.32}{0.52}
        & \bestms{73.88}{0.89}
        & \bestms{81.18}{1.08} \\
        \bottomrule
    \end{tabular}%
    }
\end{table*}

\paragraph{Properties.}
BOS Summary is training-free by construction: it introduces no additional
tokens, parameters, or auxiliary modules, and the summary is derived
entirely from the model's own hidden states and frozen attention
projections rather than injected from an external encoder. The
intervention is also confined to prefill, it modifies one attention row
at one layer, an $O(n)$ change to a prefill pass that already costs
$O(n^2)$ per layer, and once the BOS representation is formed during
prompt encoding, decoding proceeds under standard causal attention with
no further mask modification and no decoding-time overhead.

\section{Experiments}
\label{sec:experiments}

\newcommand{\bosms}[1]{\ensuremath{#1_{\pm\mathrm{TBD}}}}
\newcommand{\bestbos}[1]{\ensuremath{\mathbf{#1}_{\pm\mathrm{TBD}}}}
\newcommand{\secondbos}[1]{\ensuremath{\underline{#1}_{\pm\mathrm{TBD}}}}

\begin{table*}[!t]
    \centering
    \caption{
        Performance across four language understanding and reasoning
        benchmarks. Per-benchmark results are reported as mean Accuracy
        (\%) and F1 with standard deviation. The Average columns report
        the arithmetic mean of the benchmark scores and the arithmetic mean
        of their benchmark-wise standard deviations. The best mean within
        each model block is shown in bold, while the second-best mean is
        underlined.
    }
    \label{tab:text_results}
    \resizebox{\textwidth}{!}{%
    \begin{tabular}{lcccccccccc}
        \toprule
        \multirow{2}{*}{Method}
        & \multicolumn{2}{c}{RACE}
        & \multicolumn{2}{c}{LogiQA}
        & \multicolumn{2}{c}{MathQA}
        & \multicolumn{2}{c}{QuAC}
        & \multicolumn{2}{c}{Average} \\
        \cmidrule(lr){2-3}
        \cmidrule(lr){4-5}
        \cmidrule(lr){6-7}
        \cmidrule(lr){8-9}
        \cmidrule(lr){10-11}
        & Acc. & F1
        & Acc. & F1
        & Acc. & F1
        & Acc. & F1
        & Acc. & F1 \\
        \midrule

        \rowcolor{gray!8}
        \multicolumn{11}{c}{
            \textit{Llama-3.1-8B-Instruct}
        } \\
        \midrule

        Direct
        & \secondms{83.31}{0.85}
        & \secondms{83.13}{0.95}
        & \secondms{42.71}{5.40}
        & \meanstd{40.59}{5.82}
        & \meanstd{31.76}{1.02}
        & \meanstd{30.42}{1.20}
        & \bestms{51.60}{1.17}
        & \meanstd{17.30}{1.43}
        & \meanstd{52.35}{1.55}
        & \meanstd{42.86}{1.90} \\

        CoT
        & \meanstd{78.62}{1.11}
        & \meanstd{78.43}{1.04}
        & \meanstd{42.24}{2.51}
        & \secondms{42.13}{2.46}
        & \meanstd{61.24}{1.23}
        & \meanstd{61.32}{1.19}
        & \meanstd{42.69}{1.90}
        & \meanstd{19.20}{1.57}
        & \meanstd{56.20}{1.51}
        & \meanstd{50.27}{1.56} \\

        ZeroTuning
        & \meanstd{79.53}{1.46} & \meanstd{79.40}{1.35}
        & \meanstd{41.47}{4.96} & \meanstd{41.30}{5.04}
        & \meanstd{62.58}{1.83} & \meanstd{62.63}{1.81}
        & \meanstd{43.63}{1.46} & \meanstd{21.32}{1.38}
        & \meanstd{56.80}{2.01} & \meanstd{51.16}{2.21} \\

        SinkTrack
        & \meanstd{78.96}{1.39}
        & \meanstd{78.78}{1.36}
        & \meanstd{42.39}{4.06}
        & \meanstd{42.02}{3.86}
        & \secondms{63.48}{2.48}
        & \secondms{63.57}{2.36}
        & \meanstd{44.21}{1.42}
        & \bestms{21.72}{1.38}
        & \secondms{57.26}{1.72}
        & \secondms{51.52}{1.78} \\

        BOS Summary
        & \bestms{84.01}{0.67}
        & \bestms{83.51}{1.03}
        & \bestms{46.34}{3.15}
        & \bestms{42.33}{2.17}
        & \bestms{64.35}{1.01}
        & \bestms{64.71}{1.21}
        & \secondms{48.73}{1.91}
        & \secondms{20.33}{1.66}
        & \bestms{60.86}{1.60}
        & \bestms{52.72}{1.55} \\
        \midrule

        \rowcolor{gray!8}
        \multicolumn{11}{c}{
            \textit{Gemma-3-12B-Instruct}
        } \\
        \midrule

        Direct
        & \meanstd{86.34}{0.59}
        & \meanstd{86.17}{0.63}
        & \meanstd{58.52}{4.78}
        & \meanstd{58.02}{4.61}
        & \meanstd{32.63}{1.09}
        & \meanstd{31.11}{0.90}
        & \bestms{58.80}{1.40}
        & \bestms{34.92}{1.60}
        & \meanstd{59.07}{1.76}
        & \meanstd{52.56}{1.49} \\

        CoT
        & \meanstd{86.25}{0.94}
        & \meanstd{86.03}{1.10}
        & \meanstd{58.22}{2.51}
        & \meanstd{57.44}{2.45}
        & \meanstd{70.05}{1.97}
        & \meanstd{70.35}{1.93}
        & \meanstd{57.61}{1.00}
        & \meanstd{32.73}{1.25}
        & \meanstd{68.03}{1.61}
        & \meanstd{61.64}{1.68} \\

        ZeroTuning
        & \secondms{86.42}{0.91} & \secondms{86.22}{1.04}
        & \secondms{59.90}{2.51} & \bestms{59.34}{2.99}
        & \secondms{76.35}{1.81} & \secondms{76.31}{1.80}
        & \meanstd{54.71}{0.81} & \meanstd{30.08}{1.03}
        & \secondms{69.34}{1.34} & \secondms{62.99}{1.29} \\

        SinkTrack
        & \meanstd{86.34}{0.92}
        & \meanstd{86.11}{1.10}
        & \meanstd{59.60}{1.71}
        & \meanstd{59.07}{1.34}
        & \meanstd{76.15}{1.45}
        & \meanstd{76.15}{1.41}
        & \meanstd{54.66}{0.97}
        & \meanstd{30.11}{1.45}
        & \meanstd{69.19}{1.16}
        & \meanstd{62.86}{1.30} \\

        BOS Summary
        & \bestms{86.78}{0.94}
        & \bestms{86.22}{1.01}
        & \bestms{60.34}{1.11}
        & \secondms{59.30}{3.45}
        & \bestms{77.29}{1.41}
        & \bestms{78.14}{0.93}
        & \secondms{58.69}{1.29}
        & \secondms{34.91}{1.35}
        & \bestms{70.78}{1.14}
        & \bestms{64.64}{1.44} \\
        \bottomrule
    \end{tabular}%
    }
\end{table*}

\subsection{Experimental Setup}

\paragraph{Models.}
We evaluate BOS Summary on four instruction-tuned VLMs:
Qwen2.5-VL-7B-Instruct~\cite{bai2025qwen25vl},
Qwen3-VL-8B-Instruct~\cite{bai2025qwen3},
Gemma-3-4B-Instruct, and Gemma-3-12B-Instruct~\cite{gemmateam2025gemma3technicalreport};
and on two instruction-tuned LLMs:
Llama-3.1-8B-Instruct~\cite{grattafiori2024llama} and
Gemma-3-12B-Instruct~\cite{gemmateam2025gemma3technicalreport}.
Together, these models span multiple families, sizes, and modalities.

\paragraph{Benchmarks.}
For vision-language evaluation, we use MMBench~\cite{liu2024mmbench} and
MMStar~\cite{chen2024we} for general multimodal understanding,
M3CoT~\cite{chen2024m3cot} for multi-step multimodal reasoning, and
POPE~\cite{li2023evaluating} for object hallucination. For text-only evaluation,
we use RACE~\cite{lai2017race} for reading comprehension,
LogiQA~\cite{liu2021logiqa} for logical reasoning,
MathQA~\cite{amini2019mathqa} for mathematical reasoning, and
QuAC~\cite{choi2018quac} for conversational question answering.

\paragraph{Baselines.}
We compare BOS Summary against \textbf{Direct} prompting,
\textbf{Chain-of-Thought (CoT)}~\cite{zhang2023multimodal},
\textbf{SinkTrack}~\cite{liu2026sinktrack}, and
\textbf{ZeroTuning}~\cite{han2025zerotuning}. To ensure a controlled comparison,
all intervention-based methods are applied on top of the same CoT prompting
setup. For SinkTrack, we follow the reported configuration and apply the
intervention every five layers starting from layer~5, without additional
task-specific tuning. For ZeroTuning, we sweep the attention-scaling coefficient
over $\alpha \in \{0.5,0.9,1.0,1.1,1.5\}$. We omit
OutRo~\cite{yoo2026natureattentionsinkshapes} from quantitative comparisons
because its released pipeline targets audio-visual Omni-LLMs; applying it
to our image--text and text-only backbones would require implementation
adaptation and model-specific tuning of its rotation strength.
Baseline details and the scope of this exclusion are discussed in
Appendix~\ref{app:baselines}.

\paragraph{Metrics.}
For the multiple-choice benchmarks, we report Accuracy and Macro-F1. Accuracy is the percentage of questions answered correctly, while Macro-F1 computes F1 for each answer class and then averages across classes, reducing the effect of class imbalance. Since QuAC is an extractive question-answering benchmark, we follow its standard evaluation and report token-level F1 and exact match. Thus, the two metric columns correspond to Accuracy and Macro-F1 for multiple-choice benchmarks, and token-level F1 and exact match for QuAC. All results are averaged over three independent runs with different random seeds and reported as mean with standard deviation. Further details are provided in Appendix~\ref{app:metrics}.

\paragraph{Inference protocol.}
Direct uses the standard task-specific prompt, whereas CoT, SinkTrack,
ZeroTuning, and BOS Summary prepend the zero-shot chain-of-thought
instruction ``Let's think step by step.'' and enforce a fixed
\texttt{Answer: X} output format. For the main evaluation, all methods
decode with temperature $T=1.0$, and results are averaged over three
random seeds. Hyperparameters are selected separately for each
(model, benchmark) pair using a $10\%$ held-out calibration split. We
select the setting with the highest held-out accuracy: the intervention
layer $\ell^\star$ for BOS Summary and the attention-scaling coefficient
$\alpha$ for ZeroTuning. Final performance is measured on the remaining
$90\%$ of the data, which is not used during hyperparameter selection.
Full layer-sweep results and the selected hyperparameters for all
model--benchmark pairs are reported in
Appendix~\ref{app:hyperparameters}.

\begin{figure*}[!t]
    \centering
    \includegraphics[width=0.7\linewidth]{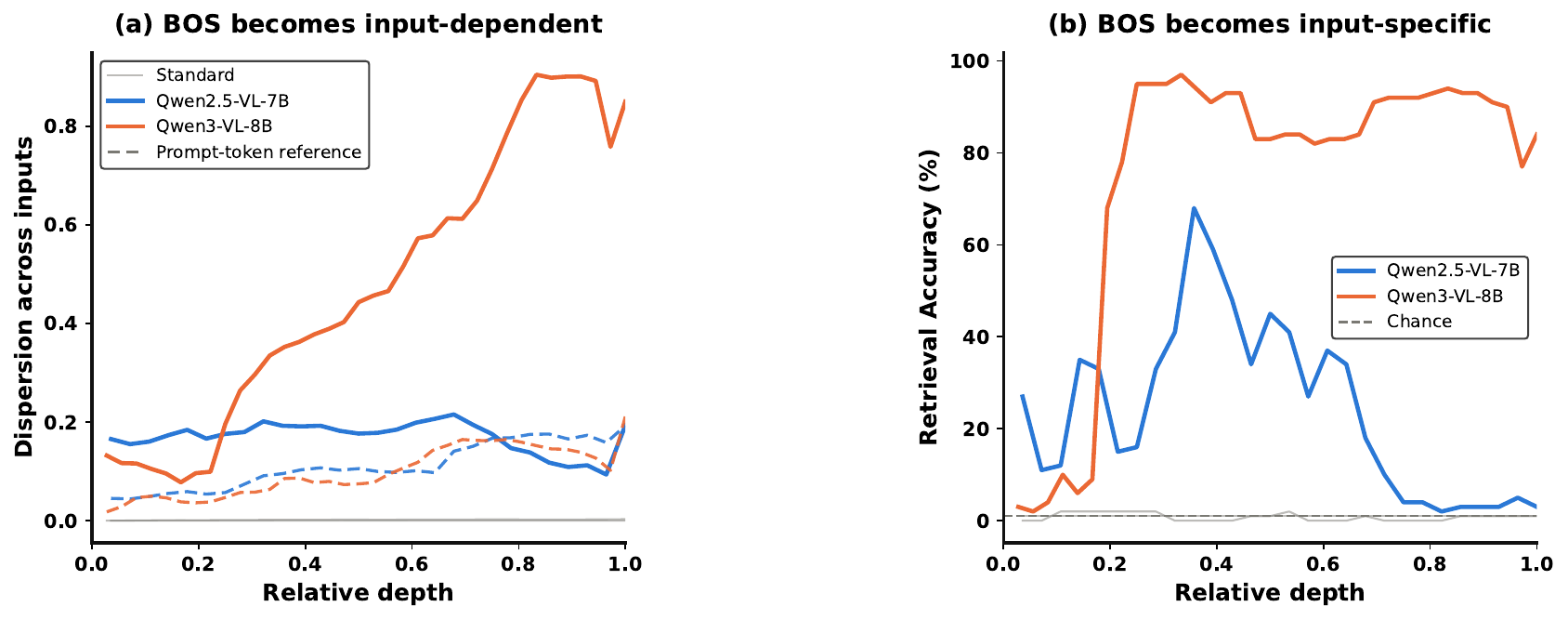}
    \caption{
\textbf{BOS Summary makes the BOS representation input-dependent and input-specific.}
Results are shown for Qwen2.5-VL-7B-Instruct and Qwen3-VL-8B-Instruct on MMStar.
(a) Dispersion measures how much the BOS representation varies across inputs.
BOS Summary increases dispersion toward the level of ordinary prompt tokens.
(b) Retrieval accuracy measures whether the BOS representation can identify its
source input. BOS Summary performs well above chance, peaking at about $68\%$
for Qwen2.5-VL-7B and $97\%$ for Qwen3-VL-8B.}
    \label{fig:bos_probe}

\end{figure*}

\subsection{Main Results}

\paragraph{Vision-language model results.}
Table~\ref{tab:main_results} shows that BOS Summary achieves the highest
Accuracy on all eight model--benchmark pairs on MMStar and M3CoT,
exceeding the strongest baseline by $0.74$--$2.58$ percentage points.
It also leads in average Accuracy for all four backbones, with gains of
$0.29$--$1.36$ points, and in average Macro-F1 for three. The improvement
extends beyond CoT prompting: BOS Summary also outperforms SinkTrack and
ZeroTuning in Accuracy on both benchmarks for every model, indicating
that the choice of sink intervention matters even under the same
prompting protocol.

The benefits vary across tasks and metrics. Qwen3-VL leads both metrics
on all four benchmarks, whereas other models show mixed results on POPE
and MMBench. On M3CoT, the Accuracy gains for Qwen2.5-VL and Gemma-3-4B
also do not yield the highest Macro-F1. These differences indicate that
improvements in overall correctness need not extend uniformly across
metrics. The consistent Accuracy gains on MMStar and M3CoT are compatible
with the proposed benefit of maintaining access to prompt evidence during
reasoning, while the mixed results elsewhere suggest that the benefit
depends on the model and task.

\paragraph{Language model results.}
Table~\ref{tab:text_results} shows that the benefit extends to text-only
inputs. BOS Summary achieves the highest values in both Average columns
for both models, exceeding the next-best method by $3.60$ and $1.20$
points on Llama-3.1-8B and by $1.44$ and $1.65$ points on Gemma-3-12B.
It attains the highest Accuracy on RACE, LogiQA, and MathQA for both
backbones. On LogiQA, Llama-3.1-8B reaches $46.34\%$, improving over
the strongest baseline by $3.63$ points. On MathQA, the Accuracy gains
over the strongest sink-intervention baseline are $0.87$ and $0.94$
points for Llama and Gemma, respectively; the corresponding Macro-F1
gains are $1.14$ and $1.83$ points.

The gains remain task-dependent. RACE improvements over the strongest
baseline are smaller, and on QuAC, Direct retains the highest token-level
F1 for both models. BOS Summary is close to Direct on Gemma's QuAC scores
but remains below it on Llama's token-level F1; SinkTrack leads Llama's
exact match. Thus, the average improvements coexist with individual
metrics that favor other methods. Together with the VLM results, this
pattern supports the usefulness of an input-conditioned BOS across
modalities, with the clearest gains on reasoning benchmarks rather than
a uniform advantage across all tasks.

\subsection{Analysis}
\label{sec:analysis}

We analyze the mechanism of BOS Summary through three questions:
\begin{enumerate}[
    label=\protect\blackcircled{\arabic*},
    leftmargin=1.8em,
    labelsep=0.45em,
    itemsep=0.2em,
    topsep=0.3em
]
    \item \emph{Does BOS become input-specific?}
    \item \emph{Does the resulting representation alter its downstream role
    as an attention sink?}
    \item \emph{Where is the displaced attention redistributed?}
\end{enumerate}
These questions trace the effect of the intervention from the BOS hidden state
to the attention patterns of subsequent layers.

\paragraph{\protect\blackcircled{1} BOS becomes input-dependent and input-specific.}
We ask whether BOS Summary makes BOS sensitive to the input. Let
$X_{i,0}^{(\ell)}$ denote BOS's hidden state for sample $i$ at layer
$\ell$, and $\bar{X}_0^{(\ell)} = \frac{1}{N}\sum_{i=1}^N X_{i,0}^{(\ell)}$
its mean over $N$ samples. We measure dispersion as
\begin{equation}
D^{(\ell)} =
\frac{\frac{1}{N}\sum_{i=1}^N \|X_{i,0}^{(\ell)}-\bar{X}_0^{(\ell)}\|_2}
     {\|\bar{X}_0^{(\ell)}\|_2}.
\label{eq:dispersion}
\end{equation}
Figure~\ref{fig:bos_probe}(a) shows the Standard BOS is nearly
input-invariant in both models. With BOS Summary, Qwen2.5-VL-7B reaches
dispersion comparable to an ordinary prompt token, and Qwen3-VL-8B
exceeds it, indicating strong input dependence.

Dispersion alone does not show BOS preserves \emph{which} input
produced it. For sample $i$ with $n_i$ tokens, define its content
signature
$S_i^{(\ell)} = \frac{1}{n_i-1}\sum_{t=1}^{n_i-1} X_{i,t}^{(\ell)}$,
the mean over non-BOS positions from the unmodified model. With
$\widetilde{(\cdot)}$ denoting mean-centering across samples, retrieval
accuracy is
\begin{equation}
R^{(\ell)} = \frac{1}{N}\sum_{i=1}^N
\mathbf{1}\big[i=\arg\max_{i'}\cos(\widetilde{X}_{i,0}^{(\ell)},\widetilde{S}_{i'}^{(\ell)})\big].
\label{eq:retrieval}
\end{equation}
Chance is $1/N$ with $N{=}500$. Figure~\ref{fig:bos_probe}(b)
shows retrieval far above chance, peaking at $68\%$ for Qwen2.5-VL-7B and $97\%$ for
Qwen3-VL-8B before declining at late layers, where the summary is progressively
transformed for downstream use: BOS Summary makes BOS both \emph{input-dependent}
and \emph{input-specific}.

\begin{figure*}[!t]
    \centering
    \includegraphics[width=0.95\linewidth]{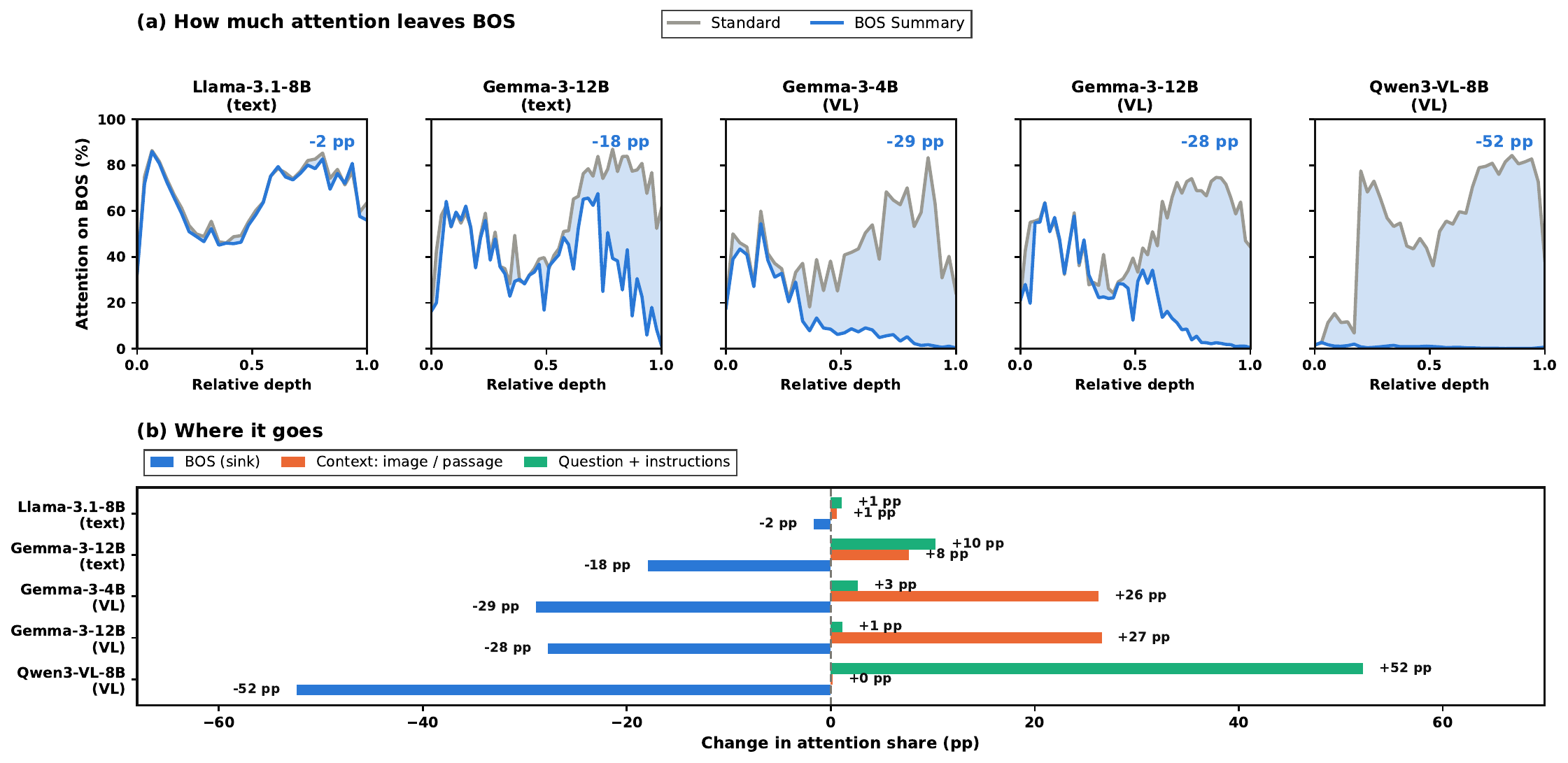}
    \caption{
\textbf{BOS Summary reduces attention to the BOS sink and redistributes it
toward the prompt.}
Results are shown for five model--modality settings using $500$ samples each.
(a)~Attention assigned to BOS across model depth for the unmodified model
(gray) and BOS Summary (blue); each panel reports the mean change.
(b)~Change in attention share for BOS, Context (image or passage), and
Question (question and instruction tokens), measured in percentage points.
All measurements use a deterministic prefill pass with eager attention.
}

    \label{fig:attention_drift}
\end{figure*}

\paragraph{\protect\blackcircled{2} The input-specific BOS weakens sink dominance.}
We next examine whether the modified BOS representation changes how later
layers attend to it. Figure~\ref{fig:attention_drift}(a) shows that BOS Summary reduces the
attention received by BOS across all evaluated model--modality settings,
although the magnitude varies by architecture. This effect is indirect:
the intervention modifies only the BOS query row at one layer, thereby
changing the resulting BOS hidden state. In subsequent layers, this modified
state changes the key and value representations associated with BOS, which
in turn affects how later tokens attend to it. Thus, a local prefill-time
modification is sufficient to weaken the downstream role of BOS as an
attention sink. This links a change in BOS content to a change in downstream
attention allocation, even though all non-BOS query rows retain their
original causal visibility.

\paragraph{\protect\blackcircled{3} Displaced attention shifts back to the input.} Finally, we examine
where the attention released from BOS is redistributed. Figure~\ref{fig:attention_drift}(b) shows the
destination is architecture-dependent: the Gemma VLMs route most released
attention to image tokens, whereas Qwen3-VL shifts it almost entirely to the
question and instruction. BOS Summary does not prescribe this destination;
the differences reflect how each pretrained model reallocates attention once
sink dominance is reduced. The shared effect is that attention moves to
content-bearing prompt tokens rather than remaining on BOS.

Taken together, the analyses suggest two complementary ways in which
BOS Summary can support access to the input. The representation probes
show that BOS carries information identifying the current prompt, while
the attention measurements show that subsequent layers allocate more
attention to content-bearing tokens. The destination of this additional
attention varies across architectures: visual context receives most of
the released attention in the Gemma VLMs, whereas the question and
instruction receive it in Qwen3-VL. A common mask modification can
therefore interact differently with each model's existing attention
patterns while improving performance across model families.

The magnitude of the attention shift alone does not explain the gains.
Llama-3.1-8B exhibits the smallest reduction in BOS attention ($2$
percentage points), yet improves its average score by $3.60$ points over
the strongest baseline. Qwen3-VL exhibits the largest reduction ($52$
points), with a smaller average score gain of $1.36$ points. This contrast suggests that the
information carried by BOS and the allocation of attention should be
considered jointly when assessing a sink intervention. Enriching BOS may
be useful even when its attention share changes little, while a larger
redistribution may help through renewed access to the prompt. These
measurements are consistent with both pathways, but do not isolate their
individual contributions to task performance.

\subsection{Fine-Tuning with BOS Summary}

\begin{table}[t]
\centering
\caption{
Fine-tuning accuracy (\%) on LogiQA ($n=651$) and RACE ($n=3498$).
$\Delta$ denotes the change in percentage points relative to the before-FT baseline.
}
\label{tab:bos_finetuning}
\small
\setlength{\tabcolsep}{3pt}
\begin{tabular}{@{}lcccc@{}}
\toprule
& \multicolumn{2}{c}{LogiQA} & \multicolumn{2}{c}{RACE} \\
\cmidrule(lr){2-3}\cmidrule(lr){4-5}
Method & Acc. & $\Delta$ (pp) & Acc. & $\Delta$ (pp) \\
\midrule
\multicolumn{5}{@{}l}{\textbf{Llama-3.1-8B}} \\
Before FT & 42.24 & -- & 83.30 & -- \\
Standard FT & 50.54 & +8.30 & 87.31 & +4.00 \\
BOS Summary FT & \textbf{53.61} & \textbf{+11.37} & \textbf{88.34} & \textbf{+5.03} \\
\midrule
\multicolumn{5}{@{}l}{\textbf{Gemma-3-12B}} \\
Before FT & 58.53 & -- & 86.34 & -- \\
Standard FT & 60.83 & +2.30 & 89.19 & +2.86 \\
BOS Summary FT & \textbf{63.44} & \textbf{+4.92} & \textbf{89.88} & \textbf{+3.54} \\
\midrule
\multicolumn{5}{@{}l}{\textbf{Gemma-3-4B}} \\
Before FT & 44.85 & -- & 73.90 & -- \\
Standard FT & 51.77 & +6.91 & 84.19 & +10.29 \\
BOS Summary FT & \textbf{52.84} & \textbf{+7.99} & \textbf{85.48} & \textbf{+11.58} \\
\bottomrule
\end{tabular}
\end{table}

\begin{figure*}[!t]
    \centering
    \includegraphics[width=0.85\linewidth]{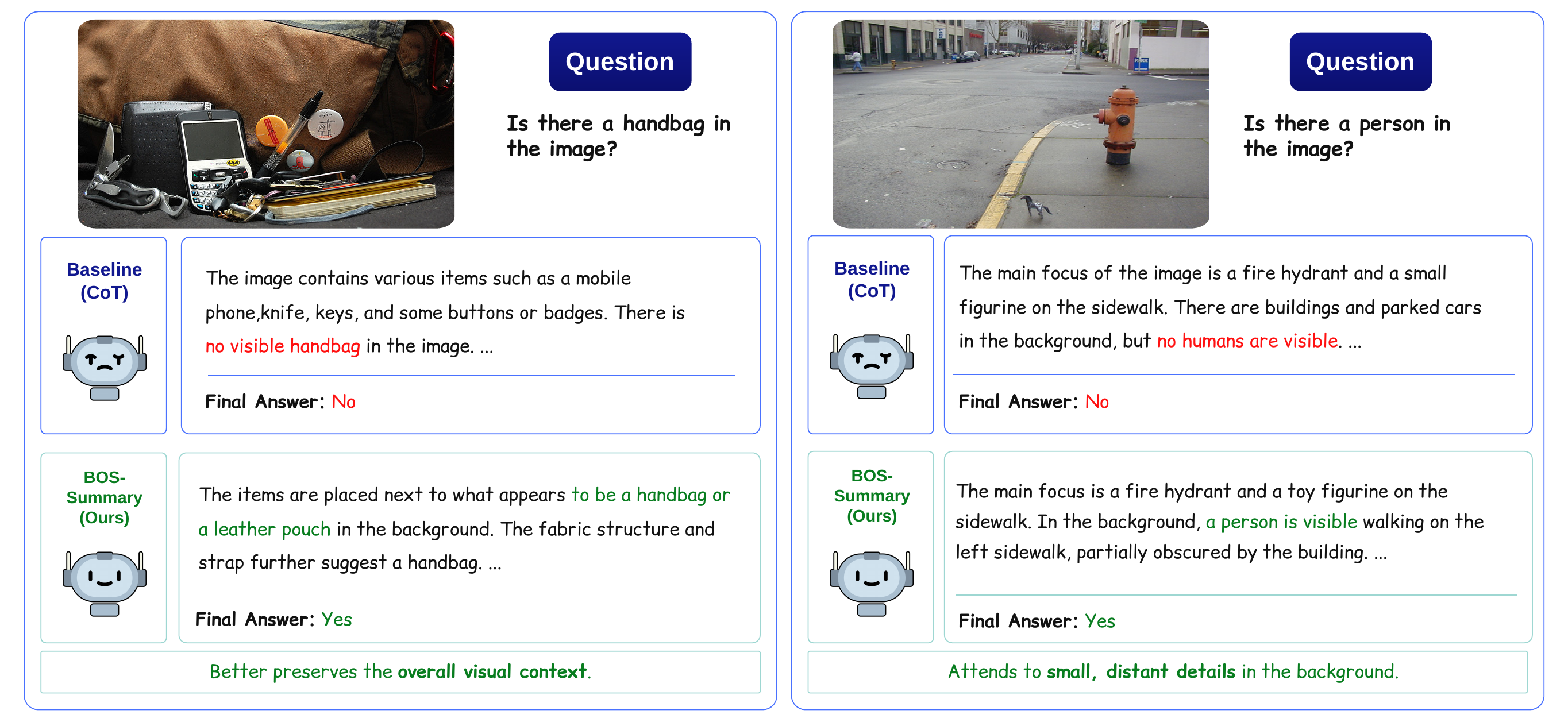}
    \caption{Qualitative comparison on two representative examples. The CoT baseline overlooks relevant visual evidence, while BOS-Summary preserves broader scene context and attends to small, distant details, leading to correct answers.}

    \label{fig:qual_figure}
\end{figure*}

We evaluate whether BOS Summary also benefits models during fine-tuning.
We compare standard LoRA fine-tuning with LoRA fine-tuning that applies
BOS Summary during both training and evaluation, using Llama-3.1-8B,
Gemma-3-12B, and Gemma-3-4B on LogiQA and RACE. Within each model--dataset
pair, both variants use the same training and evaluation settings, differing
only in the BOS Summary intervention. All results use direct prompting and
a single random seed.

Table~\ref{tab:bos_finetuning} shows that BOS Summary fine-tuning outperforms
standard fine-tuning across all six model--dataset pairs. The largest
improvement is on Llama-3.1-8B on LogiQA, where accuracy increases from
$50.54\%$ to $53.61\%$, a gain of $3.07$ percentage points. Both Gemma models
also improve on both datasets. These results indicate that the benefit of
conditioning BOS on the input extends beyond training-free inference to
fine-tuning. Under matched LoRA training, the additional gains suggest that
an input-conditioned BOS pathway can complement task-specific parameter
adaptation. The effect appears across both model families and datasets,
motivating the use of sink positions as explicit context carriers during
training. This opens a direction for learning how to aggregate and reuse
prompt information through existing attention pathways, while retaining
the backbone architecture and standard causal decoding.

\subsection{Qualitative Results}

As shown in Figure~\ref{fig:qual_figure}, BOS Summary improves grounding at both the scene and fine-grained levels. These behaviors follow from conditioning the existing BOS attention channel on the input. Because attention to BOS remains stable during decoding, the input-specific BOS helps preserve broader scene context as the generation history grows. At the same time, aggregating information across all prompt positions allows weak or peripheral evidence to remain available, helping the model recover small or partially obscured details such as the distant person in the background.

\section{Limitations}
Although BOS Summary improves standard fine-tuning, we have not analysed
how optimisation changes the BOS pathway or why this interaction yields
additional gains. Understanding this interaction could guide its use
during training and inform more transferable layer-selection rules,
beyond the current model--task-specific calibration. Establishing the scope
of these benefits also requires evaluation beyond the multiple-choice and
extractive QA tasks studied here, including open-ended visual questions
and unrestricted long-form generation.

\section{Conclusion}
We introduced \textbf{BOS Summary}, a training-free prefill-time mask edit
that turns BOS into a prompt-conditioned context carrier. Across VLMs and
LLMs, it improves reasoning and grounding while reshaping downstream
attention, and its fine-tuning gains motivate studying the same pathway
during learning. More broadly, our findings motivate research on how
attention sinks can acquire, preserve, and reuse task-relevant information.
Connecting their representational content to their downstream influence
could guide both inference-time interventions and training objectives that
use existing attention pathways to maintain access to the input during extended generation and adaptation.

\pagebreak[4]
\section*{LLM Usage Statement}
OpenAI Codex was used during manuscript preparation to assist with drafting and revising text, reviewing related literature, checking tables against experimental results, and improving document formatting and presentation. All AI-assisted content was reviewed by the authors, who remain responsible for the accuracy of the scientific claims, references, experimental results, and final manuscript.

{
    \small
    \bibliographystyle{ieeenat_fullname}
    \bibliography{main}
}

\clearpage
\appendix
\maketitlesupplementary
\section{Detailed Related Work}
\label{app:related}

\subsection{Attention Sinks in Transformers and VLMs}

The tendency of transformers to route a disproportionate share of attention to the first
token of a sequence was formalized as the \emph{attention sink} in the context of
streaming language models~\cite{xiao2024streaming}. Subsequent analysis has traced the
phenomenon to massive activations in a small subset of hidden
dimensions~\cite{sun2024massive, cancedda2024spectral}, characterized the conditions
under which it emerges during pretraining~\cite{gu2025attention}, linked it to
outlier-driven rescaling and specialized sink heads~\cite{qiu2026unified}, and argued
that it functions as an implicit no-op that regulates information mixing across
depth~\cite{barbero2025first}. Analogous
high-norm tokens observed in vision transformers have motivated explicit register
tokens~\cite{darcet2024registers}. In VLMs, sinks additionally arise on weakly informative
visual patches: \citet{kang2025see} identify visual attention sinks and analyze
which heads are image-centric, while \citet{choi2026sinks} distinguish sinks that propagate
from the vision encoder from those that emerge within the language-model layers and
quantify a trade-off between global scene-level encoding and precise local evidence.
Recent analysis further argues that the sink is not inert but encodes structured, global
information that shapes decoding~\cite{yoo2026natureattentionsinkshapes}. Together these findings frame the
sink as a plausible intervention target for multimodal grounding.

\subsection{Inference-Time Interventions on the Attention Sink}

A growing line of training-free methods intervenes on the sink at inference time. One
group rescales the attention received by the initial token: ZeroTuning~\cite{han2025zerotuning}
applies head-specific scaling to the attention that downstream tokens direct at the sink,
and ACT~\cite{yu2024unveiling} identifies and calibrates hidden non-beneficial sinks to reshape
attention distributions. A second group changes what the sink encodes.
SinkTrack~\cite{liu2026sinktrack} runs a parallel cross-attention track that injects
externally derived features (e.g., a global image embedding) into the \texttt{<BOS>}
representation, requiring auxiliary embeddings, image-position tracking, and per-layer
refresh logic. LSG~\cite{choi2026sinks} adds a lightweight gating module that dynamically
scales V-sink contributions, learned via next-token prediction with the backbone frozen.
OutRo~\cite{yoo2026natureattentionsinkshapes} lifts the causal constraint on sink positions so that the sink
query attends to the full sequence, but combines this with per-layer sink detection and a
gated head-output rotation applied throughout decoding, and reports on video and
audio-visual question answering that the mask-relaxation component alone yields only
marginal gains. Orthogonally, SinkRouter~\cite{liu2026sinkrouter} treats the sink as an
efficiency signal, skipping sink-dominated attention head computations to accelerate
long-context decoding. Our BOS Summary sits within this family as a minimal variant:
it isolates the causal-mask relaxation on \texttt{<BOS>} at prefill only, with no
auxiliary embedding, no learned parameters, no sink detection, and no head-output
rotation.

\subsection{Attention and Decoding Interventions for VLM Hallucination}

A broader landscape of training-free methods targets VLM hallucination by manipulating
attention across content tokens or by intervening on the output distribution. Within the
attention-manipulation family, PAI~\cite{liu2024paying} uniformly amplifies attention on
visual tokens, VAF~\cite{yin2025clearsight} enhances visual attention within mid-layer fusion
sites, PAINT~\cite{arif2025paint} selectively boosts image-informed local and summary
tokens, and VAR~\cite{kang2025see} redistributes attention away from visual sinks
toward content-bearing patches within image-centric heads. In the LLM setting,
PASTA~\cite{zhang2023tell} similarly reweights attention toward user-specified spans on
a small subset of heads. FastV~\cite{chen2024image} approaches visual redundancy from
the opposite direction by pruning visual tokens after the earliest layers. Beyond
attention, OPERA~\cite{huang2024opera} diagnoses over-trust on summary tokens and applies
a logit penalty with beam-search rollback, VCD~\cite{leng2024mitigating} contrasts logits against
a degraded-image counterpart, and DoLa~\cite{chuang2024dola} contrasts logits across
depth. Training-based approaches such as LRV~\cite{liu2024mitigating} instead mitigate
hallucination through robust instruction tuning, and Woodpecker~\cite{yin2024woodpecker}
takes a post-hoc correction approach that revises generated outputs against evidence
extracted from the image. Standard evaluation is carried out on hallucination-focused
benchmarks such as POPE~\cite{li2023evaluating}, AMBER~\cite{wang2023amber}, and
CHAIR~\cite{rohrbach2018chair}, as well as general capability benchmarks such as
MMBench~\cite{liu2024mmbench} and MME~\cite{fu2026mme}, with multimodal chain-of-thought
reasoning as a further evaluation setting~\cite{zhang2023multimodal, chen2024m3cot}.
In contrast to methods operating on the distribution of attention across content tokens
or on the output logits, our method modifies the \emph{content} of the sink itself, so
that its role as a broadcast channel carries a self-derived summary of the visual context
throughout generation.

\section{Visualization of the BOS Summary Mechanism}
\label{app:fig_bos_summary}
\begin{figure*}[t]
    \centering
    \includegraphics[width=0.80\linewidth]{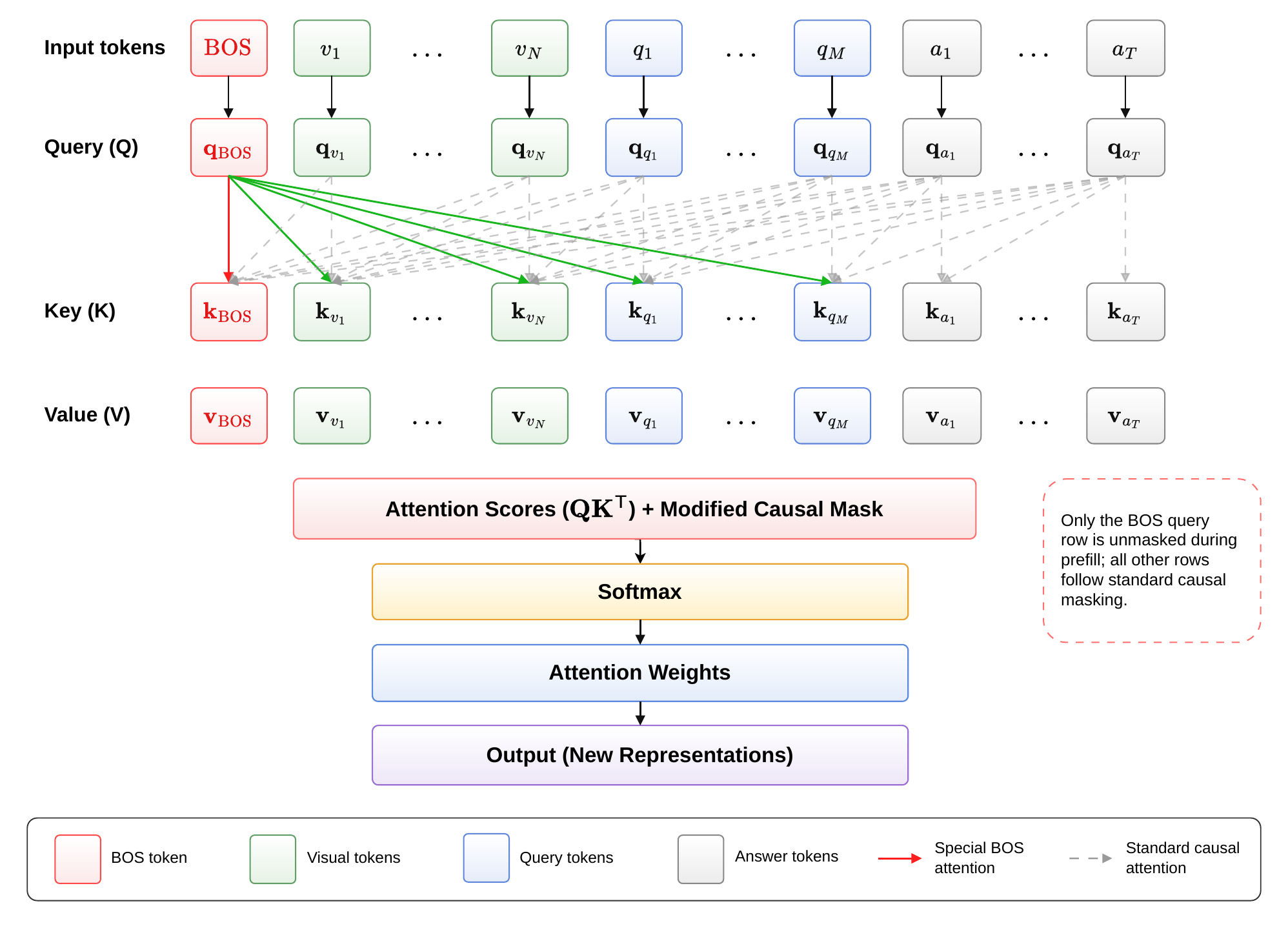}
    \caption{\textbf{Attention pattern under BOS Summary at the injection layer
$\ell^\star$ during prefill.} Each input token is projected to a query, key,
and value. Under standard causal masking (gray dashed arrows), every token
attends only to itself and to preceding positions, so the BOS query attends
solely to its own key (red arrow). BOS Summary removes the mask on the BOS
query row alone, allowing $q_{\mathrm{BOS}}$ to additionally attend to the
keys of all visual tokens $v_1,\ldots,v_N$ and query
tokens $q_1,\ldots,q_M$ (green arrows). The resulting attention output writes
a content-weighted aggregate of the full prompt into the BOS hidden state.
All other query rows are unchanged, and answer tokens $a_1,\ldots,a_T$ are
not yet present during prefill; at decoding time no mask modification is
applied.}
    \label{fig:mask_schematic}
\end{figure*}

Figure~\ref{fig:mask_schematic} illustrates the mechanism at the injection
layer. In a standard decoder-only transformer, the additive causal mask
restricts each query to its own position and those preceding it. For the
first position this leaves a degenerate case: the BOS query has no preceding
context, so its attention distribution is one-hot on itself and its output
reduces to its own value vector, independent of the input. BOS Summary
removes the mask entries on this single row, so that $q_{\mathrm{BOS}}$
attends across the full set of prompt keys. Because attention is computed in
parallel over positions, this requires no additional forward pass: the same
prefill computation now produces a BOS output that aggregates the visual and
textual prompt in proportion to the model's own attention scores. Every
other row of the mask is untouched, so all non-BOS tokens retain exactly the
causal attention pattern they had before the intervention. Answer tokens are
shown in the figure for completeness but are not present during prefill;
they are produced during decoding, at which point no mask modification is
applied and BOS is read through the frozen key and value entries already
written to the cache.

\section{Baselines}
\label{app:baselines}

\paragraph{Direct.}
The model is prompted with the task input and instructed to produce the final
answer immediately, without intermediate reasoning. This is the standard
zero-shot protocol and establishes the backbone's baseline capability under
unmodified inference. It also serves as a reference point for benchmarks where
extended reasoning is unnecessary or counterproductive, such as short yes/no
probes, on which chain-of-thought prompting can degrade performance.

\paragraph{Chain-of-Thought (CoT).}
The model is prompted to produce intermediate reasoning steps before its final
answer using the zero-shot instruction ``Let's think step by
step.''~\cite{zhang2023multimodal}. CoT is the most direct comparison for our
setting: it improves multi-step reasoning purely through prompting, without
modifying the model's computation. Because BOS Summary and the remaining
baselines are all applied on top of the same CoT prompt, CoT also isolates the
contribution of each intervention from the effect of the prompt itself.

\paragraph{ZeroTuning.}
ZeroTuning~\cite{han2025zerotuning} is a training-free intervention that
rescales the attention that downstream tokens direct at the initial token,
applying head-specific adjustments to the sink's attention logits. It represents
the family of methods that modify \emph{how much} attention the sink receives
while leaving its representation unchanged, making it the natural contrast to
our approach of changing \emph{what} the sink encodes. We sweep the scaling
coefficient over $\alpha \in \{0.5,0.9,1.0,1.1,1.5\}$ and select $\alpha$ on
the held-out calibration split.

\paragraph{SinkTrack.}
SinkTrack~\cite{liu2026sinktrack} injects externally derived contextual features
into the BOS representation through an auxiliary cross-attention track,
refreshing the anchor at multiple layers during prefill. It is the closest
baseline to BOS Summary in objective, both aim to make the sink carry input
context, but differs in mechanism: SinkTrack supplies that context from an
externally maintained embedding, whereas BOS Summary derives it within the
model's own attention. We follow the reported configuration and apply the
intervention every five layers starting from layer~5, without additional
task-specific tuning.

\paragraph{OutRo (not included in quantitative comparisons).}
OutRo~\cite{yoo2026natureattentionsinkshapes} combines sink-query mask
relaxation with gated rotation of non-sink attention-head outputs toward
the sink value direction. Its main evaluation concerns video and
audio-visual QA with Omni-LLMs. The released
implementation\footnote{\url{https://github.com/kaistmm/OutRo}}
integrates Qwen2.5-Omni and provides AVHBench and DailyOmni inference
pipelines. The paper also reports visual-only VLM experiments and
image--text and text-only generalisation, including Llama-3.1-8B.
Its rotation strength $\gamma$ is selected separately for each model;
Appendices B.5 and B.7 of that work describe the sensitivity analysis
and validation-based selection. This is inference-hyperparameter tuning,
with no model-weight updates. A comparison on our backbones and benchmarks
would require porting the released pipeline and calibrating $\gamma$
under our evaluation protocol. We leave this adaptation outside the
present quantitative comparison.

\section{Metrics}
\label{app:metrics}

This section provides further details on the metrics described in the main
text. All methods use the same prompts, decoding settings, answer-extraction
procedure, and scoring code. Reported differences therefore arise from the
inference methods rather than from changes in evaluation.

\subsection{Multiple-Choice Metrics}
\label{app:metrics_mcq}

The multiple-choice benchmarks are M3CoT, MMStar, POPE, and MMBench for the
vision-language evaluation, and RACE, LogiQA, and MathQA for the text-only
evaluation. We report Accuracy and Macro-F1 for these benchmarks.

Let $\{(\hat{y}_i,y_i)\}_{i=1}^{N}$ denote the predicted and gold labels for
$N$ questions. Accuracy is the fraction of predictions that match the
corresponding gold labels:
\begin{equation}
    \mathrm{Acc}
    =
    \frac{1}{N}
    \sum_{i=1}^{N}
    \mathbf{1}\!\left[\hat{y}_i = y_i\right].
\label{eq:accuracy}
\end{equation}

For Macro-F1, each answer option is treated as a separate class. For class
$c \in \mathcal{C}$, let $\mathrm{TP}_c$, $\mathrm{FP}_c$, and
$\mathrm{FN}_c$ denote the numbers of true positives, false positives, and
false negatives. Per-class precision, recall, and F1 are defined as
\begin{equation}
\begin{aligned}
    P_c
    &=
    \frac{\mathrm{TP}_c}
         {\mathrm{TP}_c+\mathrm{FP}_c},
    \\
    R_c
    &=
    \frac{\mathrm{TP}_c}
         {\mathrm{TP}_c+\mathrm{FN}_c},
    \\
    F_c
    &=
    \frac{2P_cR_c}
         {P_c+R_c}.
\end{aligned}
\label{eq:perclass_f1}
\end{equation}
We set $F_c=0$ when $P_c+R_c=0$. Macro-F1 is the unweighted mean of the
per-class F1 scores:
\begin{equation}
    \mathrm{Macro\text{-}F1}
    =
    \frac{1}{|\mathcal{C}|}
    \sum_{c \in \mathcal{C}} F_c.
\label{eq:macro_f1}
\end{equation}

Because all classes contribute equally, Macro-F1 is less affected by dominant
classes than Accuracy. This distinction is especially relevant for POPE, which
is evaluated as a binary yes/no classification task.

\subsection{QuAC Metrics}
\label{app:metrics_quac}

QuAC is the only extractive question-answering benchmark in our evaluation. We
follow its standard QuAC/SQuAD-style evaluation and report token-level F1 and
exact match.

Predictions and references are normalized by converting text to lowercase,
removing punctuation and the articles \emph{a}, \emph{an}, and \emph{the}, and
collapsing repeated whitespace. Token-level F1 is then computed from the
overlap between the normalized prediction and reference tokens. Let
$\mathrm{num\_same}$ denote the number of shared tokens. Precision, recall,
and F1 are
\begin{equation}
\begin{aligned}
    P
    &=
    \frac{\mathrm{num\_same}}
         {|\text{prediction tokens}|},
    \\
    R
    &=
    \frac{\mathrm{num\_same}}
         {|\text{reference tokens}|},
    \\
    F_1
    &=
    \frac{2PR}{P+R}.
\end{aligned}
\label{eq:token_f1}
\end{equation}
We set $F_1=0$ when the prediction and reference have no token overlap. Exact
match is $1$ when their normalized strings are identical and $0$ otherwise.

QuAC provides multiple reference answers for each question. We therefore take
the maximum score over all valid references. For unanswerable questions, we
follow QuAC's majority rule. When at least half of the references are
\texttt{CANNOTANSWER}, the question is treated as unanswerable and the model
must also abstain to receive credit. Otherwise, the \texttt{CANNOTANSWER}
references are removed before scoring.

For CoT outputs, the predicted span is taken from the text following
\texttt{Answer:}. Any generated continuation into the next dialogue turn is
removed, as are stray literal span-markup tags.

\subsection{Aggregation Across Seeds}
\label{app:metrics_aggregation}

\begin{figure*}[!t]
    \centering
    \includegraphics[width=\linewidth]{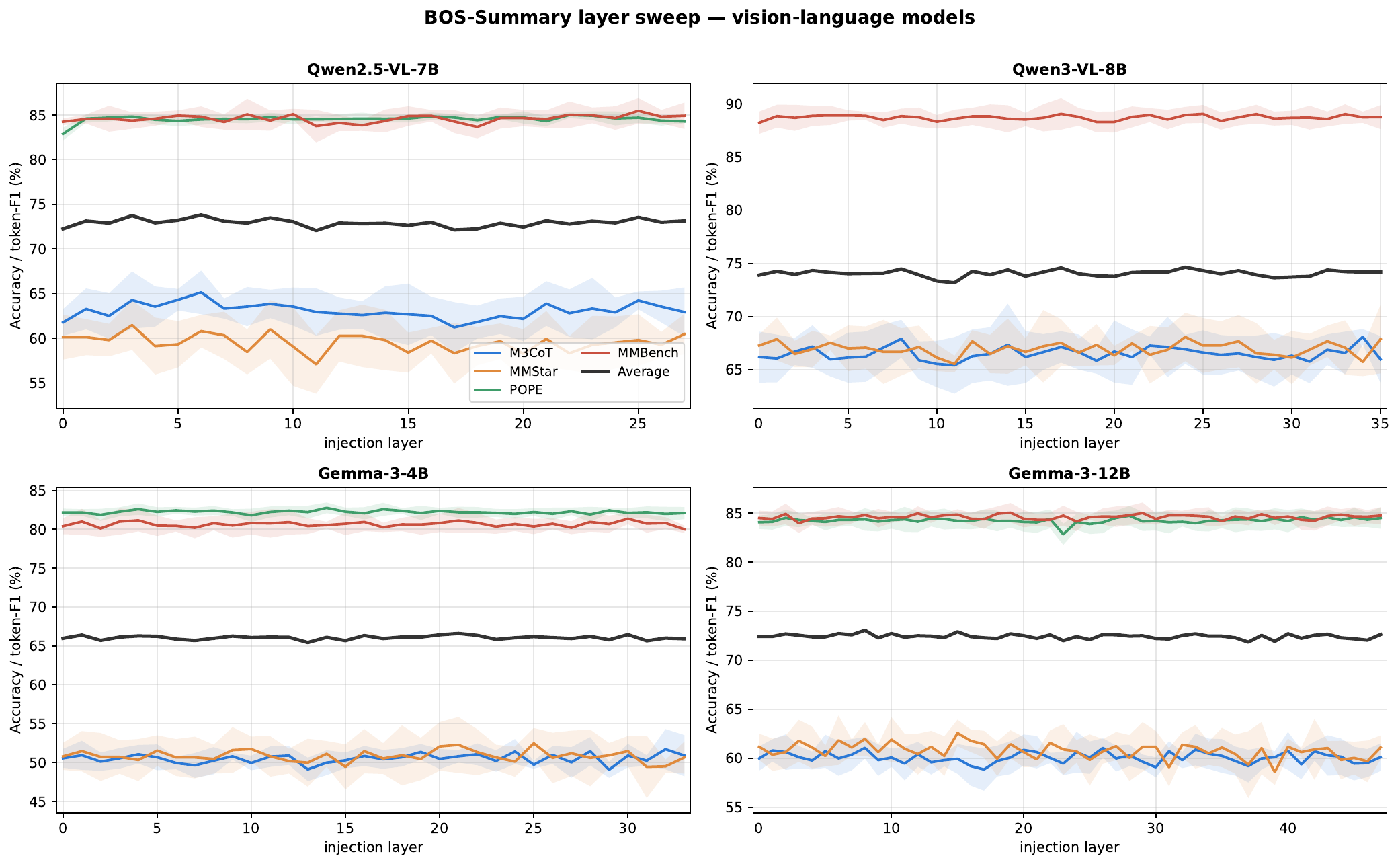}
    \caption{\textbf{VLM layer sweep results}}
    \label{fig:vlm_layer_sweep}
\end{figure*}

\begin{figure*}[!t]
    \centering
    \includegraphics[width=\linewidth]{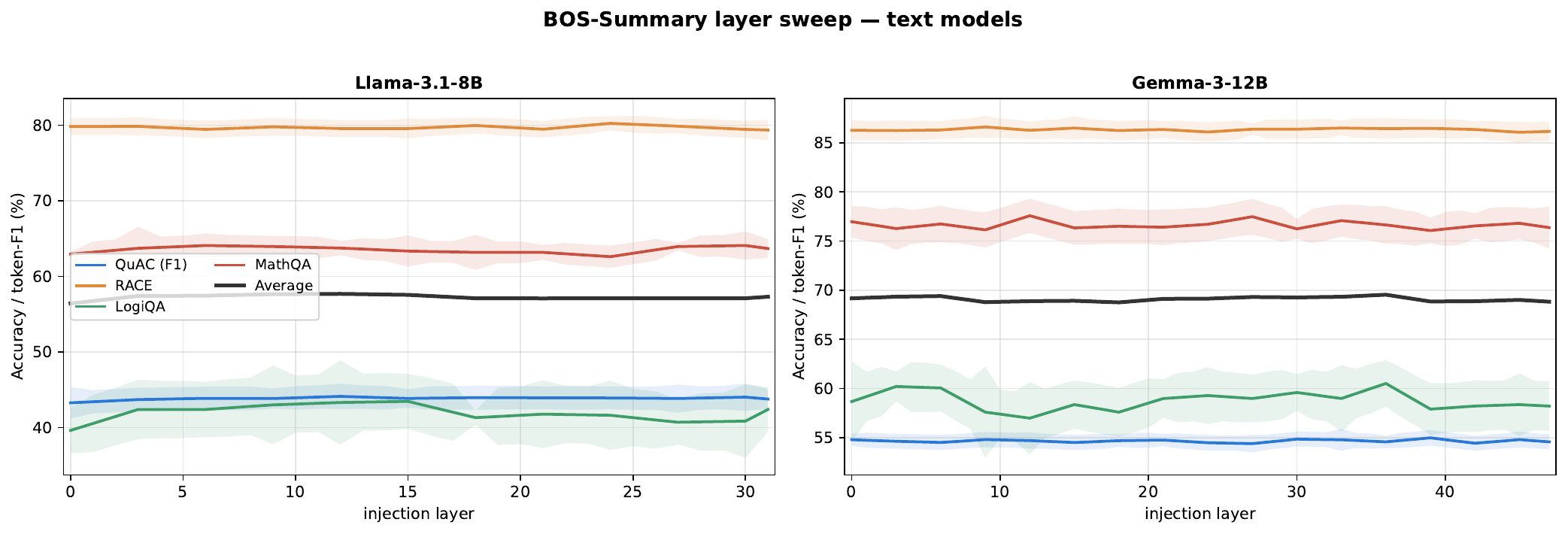}
    \caption{\textbf{LLM layer sweep results}}
    \label{fig:llm_layer_sweep}
\end{figure*}

Each reported result is averaged over three independent runs with different
random seeds. We report the mean and standard deviation across these runs. All
methods use the same test splits and evaluation code, so variation across seeds
reflects run-level differences rather than changes in scoring.

\section{Complete Results and Hyperparameters}
\label{app:hyperparameters}

We report the complete layer-wise results for every model--benchmark pair.

\subsection{Hyperparameters}
\label{app:hyperparameter_settings}

\paragraph{Models and benchmarks.}
The vision-language experiments use Qwen2.5-VL-7B,
Qwen3-VL-8B, Gemma-3-4B, and Gemma-3-12B. The text experiments use
Llama-3.1-8B and Gemma-3-12B. All models are used off the shelf, and all
inference-time interventions are training-free. The vision-language benchmarks
are M3CoT, MMStar, POPE, and MMBench. The text benchmarks are RACE, LogiQA,
MathQA, and QuAC. Accuracy and Macro-F1 are reported for the multiple-choice
tasks, including the binary yes/no POPE benchmark. QuAC is evaluated using
token-level F1 and exact match, as defined in Appendix~\ref{app:metrics}.
Every result is reported as the mean and standard deviation over three random
seeds.

\paragraph{Compared methods.}
All methods use the same task prompts, decoding settings, answer-extraction
procedure, and scoring code within each model--benchmark setting.
\textbf{Direct} answers without producing an explicit reasoning trace.
\textbf{Chain-of-Thought (CoT)} instructs the model to reason before returning
its prediction in the fixed format \texttt{Answer: X}.

\textbf{SinkTrack} intervenes at every fifth layer, namely layers
$5,10,15,20,$ and $25$ when those layers are present. At each selected layer,
the first token replaces its standard self-attention update with cross-attention
to a global representation of the input. This representation consists of the
image-token states for vision-language models and the prompt hidden states for
text models. The input representation is read again at each intervention layer
without positional encoding. The intervention is applied only during prefill
and always targets the first token.

\textbf{ZeroTuning} scales the first token's post-softmax attention weight by a
single factor $\alpha$ across all layers, heads, and decoding steps, followed by
renormalization. The identity setting is $\alpha=1$. We sweep
\[
\alpha \in \{0.5,0.9,1.0,1.1,1.5\}
\]
and report the best value separately for each model--dataset pair.

\textbf{BOS-Summary} opens the first token's attention row over all real input
tokens at the selected injection layer during prefill. After this update,
decoding follows the model's ordinary causal attention pattern. We report two
variants. The main variant uses one injection layer selected separately for
each model--dataset pair by sweeping all model layers. The
\emph{inject-all} variant applies the same intervention at every layer during
prefill. No model weights are changed in either setting.

\paragraph{Decoding settings.}
Both evaluation tracks use sampling with \texttt{do\_sample=True} and
temperature $T=1.0$. For vision-language models, generation is limited to
$1024$ new tokens for CoT prompting and $64$ for Direct prompting. For text
models, CoT generation is limited to $1024$ new tokens. Direct evaluation on
the multiple-choice text benchmarks uses a single constrained forward pass
rather than free-form generation: the next-token distribution is restricted
to the candidate labels, and the highest-scoring label is selected, following
the standard MMLU/\texttt{lm-eval-harness} protocol. Direct generation on QuAC
is limited to $48$ new tokens.

\paragraph{Fine-tuning study.}
The supplementary fine-tuning experiment trains Llama-3.1-8B on the LogiQA
training split using LoRA. The LoRA rank is $r=32$, with scaling parameter
$\alpha=2r=64$ and dropout $0.05$. LoRA adapters are applied to the
query, key, value, and output projections, as well as the gate, up, and down
projections. Training runs for four epochs with a learning rate of
$10^{-4}$, a cosine schedule, and a warmup ratio of $0.03$. The per-device
batch size is $4$, with gradient accumulation used to obtain the effective
batch size, and training uses \texttt{bfloat16}. All examples use the
direct-answer prompt format.

Two training arms use identical settings, except that BOS-Summary is active
during training in one arm. The study reports three evaluation conditions:
standard fine-tuning, BOS-Summary fine-tuning evaluated with the intervention
enabled, and the same BOS-Summary-trained weights evaluated with the
intervention disabled. Each condition is run with three seeds and reported as
mean performance with standard deviation.

\subsection{Complete Results}
\label{app:complete_results}

Figures~\ref{fig:vlm_layer_sweep} and
\ref{fig:llm_layer_sweep} report the BOS-Summary layer sweeps for the
vision-language and text models, respectively. Each curve shows Accuracy as a
function of the injection layer for a fixed model--dataset pair; QuAC uses
token-level F1 in place of Accuracy.

Across models and datasets, performance remains nearly flat over much of the
network depth. BOS-Summary is therefore largely insensitive to the exact
injection layer, even though the best layer is selected separately for the main
results. This stability is consistent with the first token acting as a stable
input-summary position throughout the network: once its attention row is opened
to the prompt, similar information can be collected across a broad range of
layers. The inject-all results provide an additional reference for whether
repeated intervention offers benefits beyond a single well-placed update.
Together, the complete tables and layer sweeps show that the reported gains do
not depend on a narrow or fragile layer choice.

\section{Dataset Prompt Templates}
\label{app:prompt-templates}
This section reports the prompts used for Direct and chain-of-thought (CoT)
inference. Placeholders enclosed in angle brackets are replaced with fields
from each example. Vision--language inputs are passed through each model's
chat template as a single user message containing the image and prompt text.
Text-only inputs are provided as raw completion prompts.

\subsection{Vision--Language Datasets}

All vision--language benchmarks use the following shared templates.

\begin{datasetprompt}{Vision--Language --- Direct}{gray}
Question: <task input>

Output Format:
Give the final answer directly using the exact format:
**Answer: <final answer>**
\end{datasetprompt}

\begin{datasetprompt}{Vision--Language --- CoT}{gray}
Question: <task input>

Let's think step by step!

Output Format:
After the reasoning, give the final answer using the exact format:
**Answer: <final answer>**
\end{datasetprompt}

The contents of \texttt{<task input>} depend on the benchmark. For M3CoT,
the question is followed by its letter-labeled answer choices. MMBench uses
the same format, with an optional hint placed before the question. MMStar
stores its answer choices directly in the question field and therefore
requires no separate option block. POPE contains a binary question without
an explicit choice list, and the final answer is either \texttt{Yes} or
\texttt{No}.

For all four datasets, the image and text prompt are provided in one user
message of the form
\texttt{[\{type:image\}, \{type:text, text:<prompt>\}]}.
Model-specific special tokens are inserted by the corresponding processor's
chat template.

\subsection{Text-Only Datasets}

RACE, LogiQA, and MathQA use a common multiple-choice structure:

\begin{datasetprompt}{Text Multiple Choice --- Direct}{gray}
<task instruction>
<context field>
Question: <question>
<letter-labeled choices>
Answer:
\end{datasetprompt}

\begin{datasetprompt}{Text Multiple Choice --- CoT}{gray}
<task instruction>
<context field>
Question: <question>
<letter-labeled choices>
Let's think step by step. Briefly explain your reasoning, then on a new line
give the final answer in the exact format 'Answer: X'.
\end{datasetprompt}

For RACE, the context field is introduced as \texttt{Article:}; for LogiQA,
it is introduced as \texttt{Passage:}; and for MathQA, the input begins with
\texttt{Problem:} and does not contain a separate passage. RACE and LogiQA
use options A--D, while MathQA uses options A--E.

QuAC uses a separate extractive question-answering template:

\begin{datasetprompt}{QuAC --- Direct}{quacColor}
Answer the question using a span from the passage. If the passage does not
contain an answer, reply exactly CANNOTANSWER.

Passage: <passage>
Conversation so far:
<previous question--answer pairs>
Question: <current question>
Answer:
\end{datasetprompt}

\begin{datasetprompt}{QuAC --- CoT}{quacColor}
Answer the question using a span from the passage. If the passage does not
contain an answer, reply exactly CANNOTANSWER.

Passage: <passage>
Conversation so far:
<previous question--answer pairs>
Question: <current question>
Let's think step by step. Briefly explain your reasoning, then write
'Answer: ' followed by a span copied from the passage, or
'Answer: CANNOTANSWER'.
\end{datasetprompt}

The conversation-history block is omitted for the first turn of each QuAC
dialogue. For later turns, it contains all preceding question--answer pairs
in chronological order.

\end{document}